\documentclass[
]{ceurart}

\usepackage{listings}
\begin{document}
\copyrightyear{2023}
\copyrightclause{Copyright for this paper by its authors.
  Use permitted under Creative Commons License Attribution 4.0
  International (CC BY 4.0).}
\conference{Second International Workshop on Linked Data-driven Resilience Research (D2R2'23) co-located with ESWC 2023, May 28th, 2023, Hersonissos, Greece}
\title{A Knowledge Graph Perspective on Supply Chain Resilience}
\author[1,4]{Yushan Liu}[%
email=yushan.liu@siemens.com
]
\cormark[1]
\address[1]{Siemens AG, Otto-Hahn-Ring 6, 81739 Munich, Germany}
\address[2]{Siemens AG, Östliche Rheinbrückenstraße 50, 76187 Karlsruhe, Germany}
\address[3]{Siemens Schweiz AG, Theilerstraße 1a, 6300 Zug, Switzerland}
\address[4]{Ludwig-Maximilians-Universität München, Geschwister-Scholl-Platz 1, 80539 Munich, Germany}

\author[1,4]{Bailan He}[%
]

\author[1]{Marcel Hildebrandt}[%
]

\author[1]{Maximilian Buchner}[%
]

\author[1]{Daniela Inzko}[%
]

\author[3]{Roger Wernert}[%
]

\author[2]{Emanuel Weigel}[%
]

\author[1]{Dagmar Beyer}[%
]

\author[1]{Martin Berbalk}[%
]

\author[1,4]{Volker Tresp}[%
]
\cortext[1]{Corresponding author.}
\begin{abstract}
Global crises and regulatory developments require increased supply chain transparency and resilience. Companies do not only need to react to a dynamic environment but have to act proactively and implement measures to prevent production delays and reduce risks in the supply chains. However, information about supply chains, especially at the deeper levels, is often intransparent and incomplete, making it difficult to obtain precise predictions about prospective risks. By connecting different data sources, we model the supply network as a knowledge graph and achieve transparency up to tier-3 suppliers. To predict missing information in the graph, we apply state-of-the-art knowledge graph completion methods and attain a mean reciprocal rank of 0.4377 with the best model. Further, we apply graph analysis algorithms to identify critical entities in the supply network, supporting supply chain managers in automated risk identification.
\end{abstract}
\maketitle

\section{Introduction}
Global crises such as pandemics, natural disasters, and economic events as well as political and regulatory developments 
lead to increasing requirements regarding supply chain transparency and resilience. To ensure smooth procurement and production processes, it is essential for companies to react timely and flexibly to dynamic conditions and incidents to prevent production delays and bottlenecks within the supply network. 

Usually, only direct (tier-1) suppliers of a company are tracked in supply chain management tools. The knowledge of subsuppliers is often limited and disregarded for decision making.  In a survey, almost 80\% of the companies cannot even name the number of their tier-$n$ ($n\geq 2$) suppliers~\cite{brylowski2021studie}, let alone their names and locations. 
Intransparent supply chains make it highly challenging to achieve precise forecasts and react in the best way in case of disruptions.

Besides the intransparency of supply chains, another challenge is posed by the decentralized storage of relevant data and their incompleteness. The data come from different sources and are stored in various formats and locations. The disconnectedness makes it difficult to get a good overview of the situation and available information. Some information is also generally hard to retrieve, e.\,g., the exact production location of a material. Even if the supplier that delivers the material is known, the exact production site is often unknown.

Supply chain management involves monitoring supply chains to ensure their operability. Due to the inherent domain complexity and the high volume of data, significant blind spots at deeper levels of the supply chains remain, which matters because many of today’s most pressing supply shortages (e.\,g., in the semiconductor industry) happen at these deeper tiers. Therefore, possible risks in the supply chains need to be identified early, i.\,e., constellations in the supply chains that lack the resistance to withstand disruptive events. 
For example, constellations can be critical if many suppliers are located in the same region, multiple tier-1 suppliers buy from the same subsupplier, only one supplier is related to a specific business scope, etc. After identifying possible criticalities, strategic decisions and mitigation measures can be derived within the organization.
Risk identification is often based on domain knowledge and manual efforts. 75\% of the companies in a survey see a need for improvement with respect to risk identification methods, where the potential of machine learning approaches is valued highly~\cite{brylowski2021studie}.

In this paper, we aim at increasing supply chain resilience, based on data from Siemens, by addressing the challenges mentioned above in the following ways:
\begin{itemize}
\item Supply chain intransparency and data disconnectedness: We collect and connect supply chain-related data from different sources and create a knowledge graph, which contains information from Siemens suppliers up to tier 3. 
\item Data incompleteness: We apply state-of-the-art knowledge graph completion methods for link prediction in the knowledge graph to predict missing information.
\item Identification of criticalities: We use graph analysis algorithms to identify critical entities in the supply network, where we focus on centrality measures to derive an importance score for each supplier.
\end{itemize}

The remainder of this paper is organized as follows. Section~\ref{sec:related} outlines related work, and Section~\ref{sec:dataset} describes the supply chain knowledge graph. In Section~\ref{sec:knowledge}, we apply knowledge graph completion methods to the data, while in Section~\ref{sec:graph}, we use graph analytics to find criticalities in the supply chains. The conclusion and further research directions follow in Section~\ref{sec:conclusion}.

\section{Related Work}
\label{sec:related}
The application of machine learning for supply chain management is becoming an increasingly active field of research~\cite{brylowski2021review}. While many supervised machine learning methods (e.g., decision trees, support vector machines, and neural networks) were successfully applied to tasks related to supply chain design, planning, and execution~\cite{brylowski2021review, tirkolaee2021}, not many works exist in the area of knowledge graphs and graph machine learning. 

In 2018, \citet{brintrup2018} published the first work to apply link prediction to a supply network. They modeled the supply network as a homogeneous graph (i.\,e., containing one relation type) with handcrafted embeddings and defined a binary classification task to predict new links in the graph. A follow-up work used graph neural networks to predict the supplier relationship between companies~\cite{kosasih2021}. 
\citet{gopal2021} also used graph neural networks to predict new supplier relationships, where they included external information about companies, e.\,g., industry classification and revenue segmentation, as features. \citet{lu2020} discovered potential partnerships between companies based on graph projections and connectivity patterns. \citet{aziz2021} represented supply networks as heterogeneous graphs (i.\,e., containing several relation types, also commonly referred to as knowledge graphs) and applied a relational graph convolutional network for link prediction. 

\section{Knowledge Graph Dataset}
\label{sec:dataset}
The supply chain knowledge graph is constructed from both Siemens-internal and external sources to reflect both internal knowledge such as tier-1 suppliers, business scopes, and Siemens parts and external knowledge such as public data about smelters and substances. The information about tier-2 and tier-3 suppliers of Siemens is obtained mainly from public customs data, and a small part is obtained from private customs data and public media. There are in total 16,910 tier-1, 43,759 tier-2, and 49,775 tier-3 suppliers of Siemens, where the suppliers at different tier levels are not mutually exclusive. The graph is modeled via the graph data platform Neo4j.
 \begin{table}[b]
\caption{Entity and relation type statistics. In the graph, there are 8 entity types, where most nodes are suppliers, and 11 relation types, where most edges are from the type \textit{supplies\textunderscore to}.}
\begin{center}
\addtolength{\tabcolsep}{2pt}
\resizebox{0.63\columnwidth}{!}{
\begin{tabular}{|c|c||c|c|}
\hline
Entity type & Nodes & Relation type & Edges\\
\hline
Supplier & 61,234 & supplies\textunderscore to & 138,197 \\
Manufacturer Part & 1,650 & related\textunderscore to & 59,894 \\
Siemens Part & 1,295 & belongs\textunderscore to & 56,663 \\
Smelter & 340 & located\textunderscore in & 30,107 \\
Substance & 321 & includes & 10,088  \\
Component & 233 & produces & 7,831 \\
Country & 172 & produced\textunderscore in & 4,381 \\
Business Scope & 32 & same\textunderscore as & 1,847 \\
& & manufactured\textunderscore by & 1,564 \\
& & contains & 764 \\
& & refines & 340 \\
\hline 
Total & 65,277 & Total & 311,676 \\
\hline
\end{tabular}
}
\label{tab:entity_relation_type}
\end{center}
\end{table}

We define a knowledge graph (KG) as a collection of triples $\mathcal{G} \subset \mathcal{E} \times \mathcal{R} \times \mathcal{E} $, where $\mathcal{E}$ denotes the set of entities and $\mathcal{R}$ the set of relation types. 
 Elements in $\mathcal{E}$ correspond to supply chain-related entities, e.\,g., suppliers, smelters, and components, and are represented as nodes in the graph. Every entity has a unique entity type, which is defined by the mapping $t: \mathcal{E} \rightarrow \mathcal{T}$, where $\mathcal{T}$ stands for the set of entity types.
The entities are connected via relation types specified in $\mathcal{R}$, represented as directed edges in the graph. All entity and relation types and corresponding numbers of nodes and edges are listed in Table~\ref{tab:entity_relation_type}.

\begin{figure}
  \centering
  \includegraphics[width=0.84\linewidth]{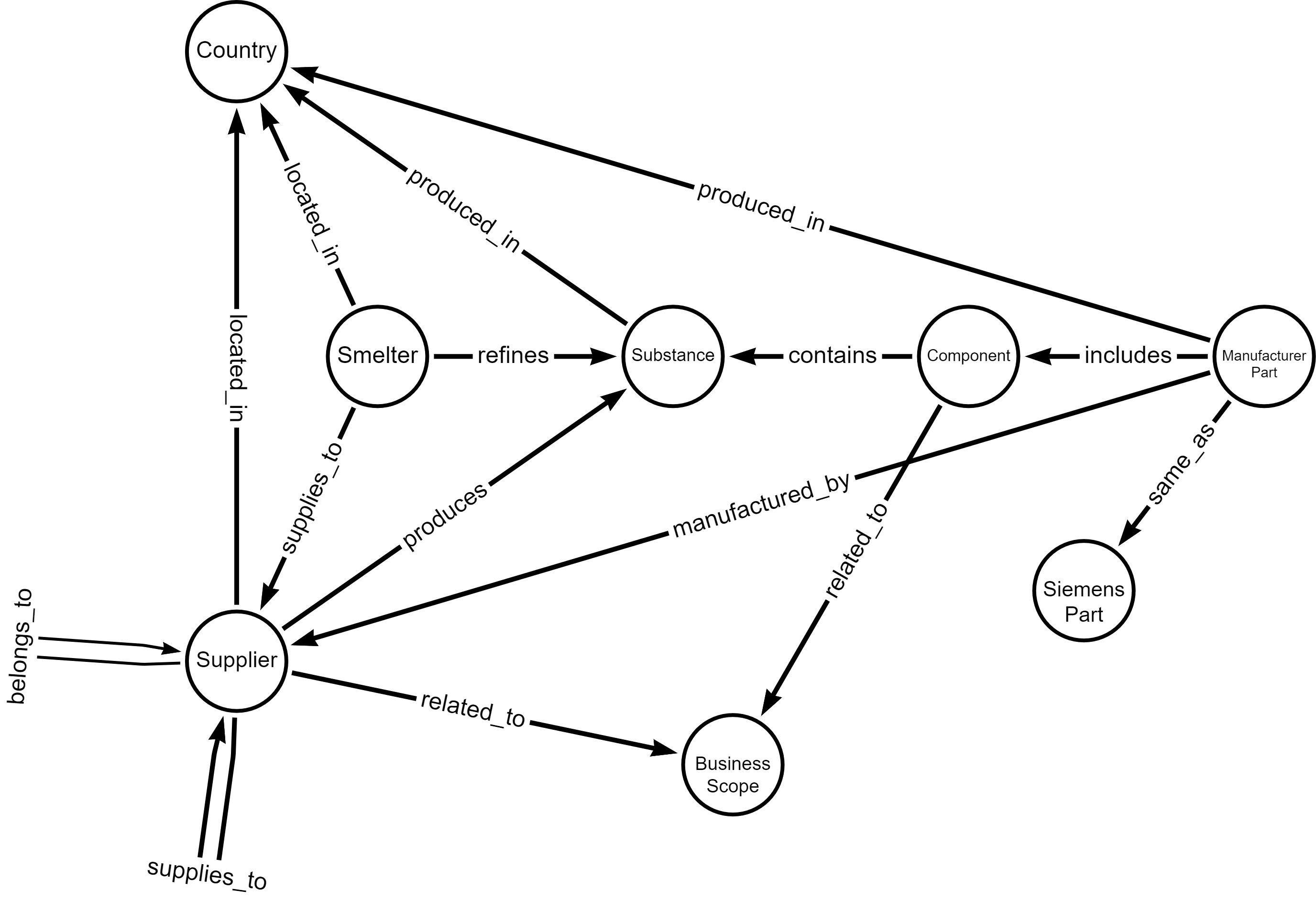}
  \caption{Knowledge graph schema. There are 8 entity types and 11 relation types.}
  \label{fig:schema}
\end{figure}

Each relation type $r \in \mathcal{R}$ connects entities from a fixed set of source entity types $\mathcal{T}_{source}(r)$ to a fixed set of target entity types $\mathcal{T}_{target}(r)$. For example, $\mathcal{T}_{source}(\mathrm{supplies\textunderscore to}) = \{\mathrm{Supplier},\, \mathrm{Smelter}\}$ and $\mathcal{T}_{target}(\mathrm{supplies\textunderscore to}) = \{\mathrm{Supplier}\}$. The schema of the graph depicts the possible connections between the entity types and is shown in Figure~\ref{fig:schema}.

A fact from the graph is represented by a triple $(\mathrm{subject}, \mathrm{predicate}, \mathrm{object}) \in \mathcal{G}$, where the subject and object are entities and the predicate is the relation type that connects them, directed at the object. Triples in the graph are assumed to be true facts, while the truth value of non-existing triples could either be wrong or unknown (since the data are highly incomplete).

\section{Knowledge Graph Completion}
\label{sec:knowledge}

\subsection{Object prediction task}
Many KGs suffer from incompleteness, so a common reasoning task in graph machine learning is KG completion or link prediction. We formulate the link prediction problem as an object prediction task. Given a query of the form $(\mathrm{subject}, \mathrm{predicate}, \mathrm{?})$, the goal is to predict a ranked list of entity candidates that are most likely the correct object of the query. 

To measure the quality of the predictions, the mean reciprocal rank (MRR) and hits@$k$ for $k \in \mathbb{N}$ are standard metrics used for link prediction on KGs. For a rank $x \in \mathbb{N}$, i.\,e., the position in the ranked list of entity candidates, the reciprocal rank is defined as $\frac{1}{x}$, and the MRR is the average of all reciprocal ranks of the correct query objects over all queries.  
The metric hits@$k$ represents the proportion of queries for which the correct object appears under the top $k$ candidates.

\subsection{Knowledge graph completion methods}
There exists a variety of methods for KG completion~\cite{wang2021}. In this paper, we focus on graph representation learning, where the underlying idea is to learn low-dimensional embeddings (i.\,e., vectors or matrices) for the entities and relation types in the graph that capture their semantic meanings. Based on these embeddings, a score can be calculated for each entity, indicating its likelihood to be the correct object of a query.

We apply the following traditional and state-of-the-art methods to the supply chain KG:
\begin{itemize}
    \item RESCAL~\cite{nickel2011} was the first method to be published for learning KG embeddings. It models the KG as a three-way tensor and performs tensor factorization for relational learning tasks such as link prediction.
    \item ComplEx~\cite{trouillon2016} was the first KG embedding method that learns embeddings in the complex vector space. It is based on tensor factorization and the Hermitian dot product.
    \item TuckER~\cite{balazevic2019} is a tensor factorization method based on the Tucker decomposition. It can be seen as a generalized version of RESCAL and ComplEx.
    \item TransE~\cite{bordes2013} was the first translational method, which models the relations between two entities as translations in the vector space.
    \item RotatE~\cite{sun2019} is a roto-translational method, which models the relations between two entities as rotations in the complex vector space.
    \item ConvE~\cite{dettmers2018} was the first method that uses convolutional neural networks to model the interactions between entities.
    \item RGCN~\cite{schlichtkrull2018} consists of a relational graph convolutional network for encoding the entities and a tensor factorization method for scoring.
    \item CompGCN~\cite{vashishth2020} incorporates composition operators to learn joint embeddings for entities and relation types. It is a generalized version of RGCN.
\end{itemize}

\subsection{Experimental setup}
For all KG completion methods, we use the implementations provided by the Python library PyKEEN~\cite{ali2021}. 
We split the graph into training, validation, and test dataset, where we operate under the transductive setting, i.\,e., all entities and relation types from the validation and test set are also included in the training set. The training set consists of 65,277 nodes and 249,340 triples, while both the validation and test set have 31,168 triples. The number of nodes for the validation and test set is 22,212 and 22,213, respectively. All three datasets include all entity and relation types. 
We use the optimizer Adam and optimize the margin ranking loss with a margin of 1, where one negative triple is sampled for each training triple. We tune the hyperparameters embedding size in the range $\{16, 32, 64, 256, 512, 1024\}$ and learning rate in the range $\{0.0001, 0.001, 0.01\}$. The training of the model is stopped early if there is no improvement regarding the metric hits@10 on three subsequent evaluations on the validation set, where the evaluation takes place every 10 epochs. 

\subsection{Results}
Table~\ref{tab:results} displays the results of the best models for the selected KG completion methods. All models were able to learn useful embeddings from the training set, while RotatE performed best with respect to all metrics. For more than $36\%$ of the test queries, RotatE predicts the correct object as the highest-ranked entity in the candidates list, and for almost half of the queries, RotatE is able to predict the correct object under the top three entities. Out of the three tensor factorization methods (RESCAL, ComplEx, and TuckER), ComplEx, which learns embeddings in the complex vector space, performs best. Out of the three neural network-based methods (ConvE, RGCN, and CompGCN), RGCN performs best, while ConvE and CompGCN have similar performance. Before materializing the results in the graph, domain experts should check the predicted triples for plausibility.
\begin{table}
\caption{Results on the test dataset for the object prediction task. The best results are displayed in bold.}
\begin{center}
\addtolength{\tabcolsep}{2pt}
\begin{tabular}{|c | c c c c|}
\hline
Method & MRR & Hits@1 & Hits@3 & Hits@10\\
\hline
RESCAL & 0.1476 & 0.0684& 0.1809 & 0.2772 \\
ComplEx & 0.2535 & 0.1793 & 0.2850 & 0.3949\\
TuckER & 0.1738 & 0.0749 & 0.1878 & 0.4033\\
TransE & 0.1595 & 0.0873& 0.1733 & 0.3164 \\
RotatE & \textbf{0.4377} & \textbf{0.3686} & \textbf{0.4733} & \textbf{0.5627} \\
ConvE & 0.2289 & 0.1549 & 0.2438 & 0.3875 \\
RGCN & 0.2911 & 0.1784 & 0.3379 & 0.5195\\
CompGCN & 0.2223 & 0.1271 & 0.2486 & 0.4229\\
\hline
\end{tabular}
\label{tab:results}
\end{center}
\end{table}

\begin{figure}
  \centering
  \includegraphics[width=\linewidth]{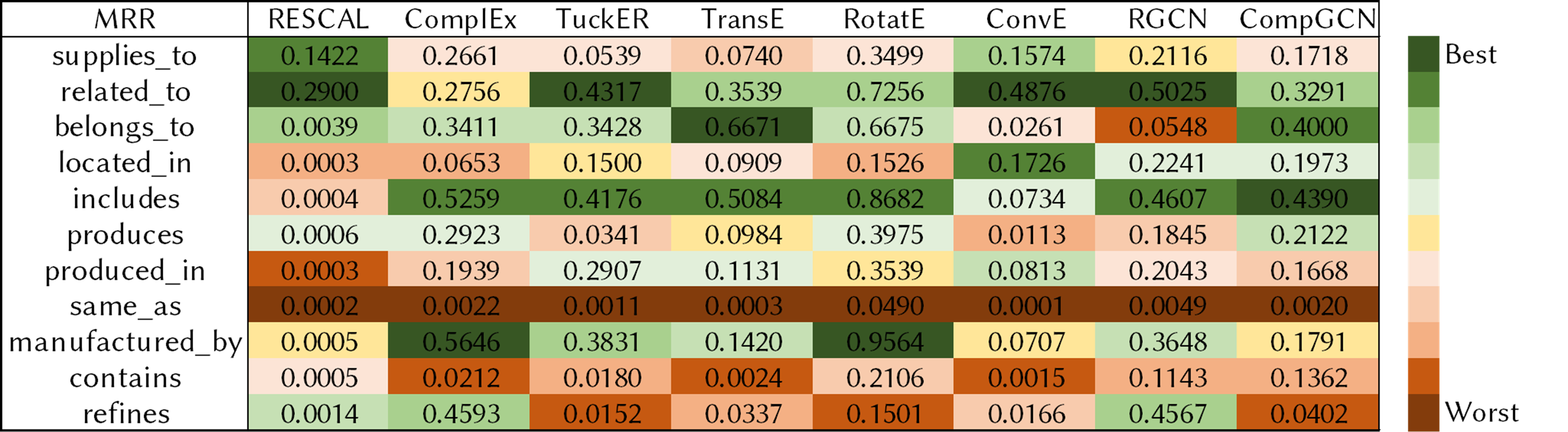}
  \caption{Results of the best models for each relation type. For each model, the performance with respect to the MRR is colored from best (green) to worst (red).}
  \label{fig:relations}
\end{figure}

In Figure~\ref{fig:relations}, the results of the best models are shown for each relation type. For every model, the performance with respect to the MRR is colored from best (green) to worst (red). RotatE performs best for all relation types except for \textit{locate\textunderscore in} (where all neural network-based models have higher MRR) and \textit{refines} (where ComplEx and RGCN are better). The models show varying degrees of performance for the different relation types, where most models tend to have good results on \textit{related\textunderscore to}, \textit{includes}, and \textit{belongs\textunderscore to} and bad performance on \textit{same\textunderscore as}, \textit{contains}, and \textit{refines}. Especially \textit{same\textunderscore as} is worst for all models. The reason lies in the graph schema and structure, where the \textit{same\textunderscore as} relation type connects manufacturer parts and Siemens parts, which do not have any other connections. Without further information, it is difficult to predict the correct Siemens part as query object.  

\section{Graph Analytics}
\label{sec:graph}
\subsection{Graph analysis algorithms}
Supply chain managers mainly decide based on tier-1 supplier data whether there are risks in the supply chains, and they are usually directly informed by tier-1 suppliers if there are already existing problems. If additional data about tier-$n$ suppliers are available, more detailed and precise decisions can be made, and mitigation measures in an earlier phase of the supply chain can be enabled. However, this kind of approach is mainly reactive, and manual decision making is not scalable to complex supply networks. Therefore, we propose to use graph analytics to support supply chain managers by automatically identifying critical suppliers so that they can be monitored more closely and mitigation strategies can be derived together. 

For the graph analysis, we use the Neo4j Graph Data Science library and concentrate on the subgraph consisting of supplier entities and the relation type \textit{supplies\textunderscore to}. We calculate the following centrality and community detection metrics, which serve as a basis for deriving the importance or criticality of a supplier:
\begin{itemize}
    \item The   \textbf{degree centrality} measures the number of incoming and outgoing edges for each node. The number of incoming edges represents the number of suppliers and the number of outgoing edges the number of customers for each company. Companies with high in- or out-degree might be affected by disruptive events more often. 
    \item The \textbf{betweenness centrality} for each node is based on the number of shortest paths between all node pairs that the node lies on. A company with high betweenness connects many companies and is more likely to cause a bottleneck. 
    \item The \textbf{closeness centrality} measures the average length of the shortest paths between a node and all other nodes. A company with high closeness is a central customer for many suppliers.
    \item The \textbf{triangle count} is a community detection measure that calculates the number of adjacent triangles of a node. A company with a high triangle count is part of an interconnected supply network. 
\end{itemize}
To make the suppliers better comparable, we normalize the metrics in-degree, out-degree, betweenness, closeness, and triangle count to be between 0 and 10 and sum them up to obtain an aggregated importance score for each supplier in the graph.

\subsection{Results}
\begin{table}
\caption{Correlation matrix. There is a high correlation between in-degree, betweenness, and triangle count.}
\begin{center}
\addtolength{\tabcolsep}{0pt}
\begin{tabular}{|c | c c c c c|}
\hline
Correlation & in-degree & out-degree & betweenness & closeness & triangle count\\
\hline
in-degree & 1.0000 & & &  &\\
out-degree & 0.1969 &1.0000&  &  & \\
betweenness & 0.8816 & 0.3928 & 1.0000 & &\\
closeness & 0.0686 & 0.2792& 0.0859 & 1.0000 &\\
triangle count & 0.9809 & 0.2048 & 0.8774 & 0.0542 & 1.0000 \\
\hline
\end{tabular}
\label{tab:correlation}
\end{center}
\end{table}
When comparing the aggregated scores of the suppliers, Siemens is obviously the center of the supply network and has an aggregated score of 37.25. The next supplier has a score of only 16.66, and there are only 3 suppliers with a score above 15. There are in total 988 suppliers with a score above 10, which might be critical entities in the supply network and should be examined by domain experts. Since an aggregated score loses information, the suppliers with the highest values for each metric should be analyzed in more detail.
Table~\ref{tab:correlation} shows the correlation matrix of the five metrics. There is a high correlation between in-degree, betweenness, and triangle count. That means, companies with many suppliers often lie on a large number of shortest paths (supply chains) and are part of a highly interconnected supply network.

\begin{figure}
  \centering
  \includegraphics[width=0.77\linewidth]{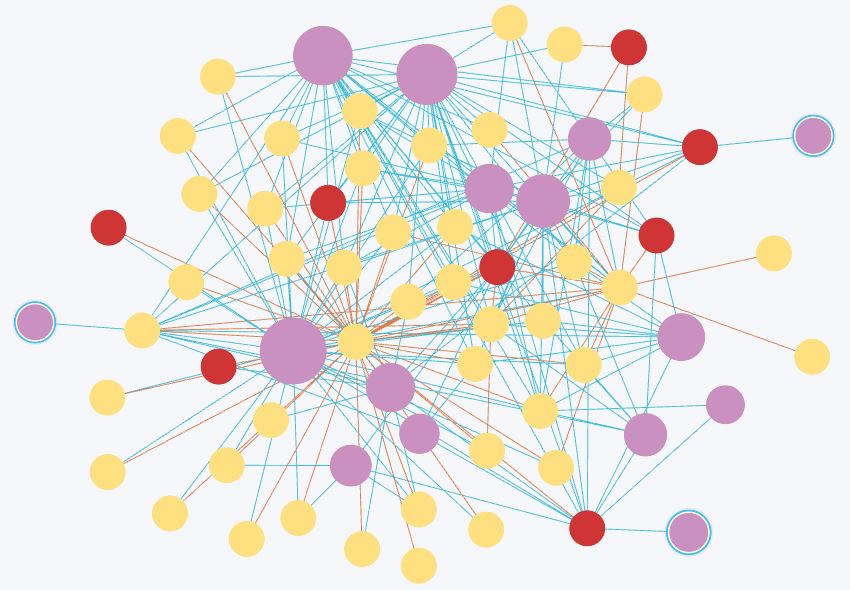}
  \caption{Visualization of a subgraph. The yellow nodes and red nodes represent suppliers, where the red suppliers have aggregated scores above 10. Business scopes are represented as purple nodes. The size of the purple nodes correlates to the number of suppliers related to the corresponding business scopes. The orange edges represent the edge type \textit{supplies\textunderscore to} and the blue edges the edge type \textit{related\textunderscore to}.}
  \label{fig:visualization}
\end{figure}

Figure~\ref{fig:visualization} illustrates a possible way to visualize the results in order to identify critical paths in the supply network. The subgraph contains yellow and red nodes, which represent suppliers, and purple nodes, which represent business scopes. The red suppliers have aggregated scores above 10 and might be more critical than the yellow suppliers. Any supply chain containing a critical supplier might have a higher risk. The size of the purple nodes corresponds to the number of suppliers related to the corresponding business scopes. The orange edges represent the edge type \textit{supplies\textunderscore to} and the blue edges the edge type \textit{related\textunderscore to}. If there are business scopes to which only one supplier is related, then the supplier might be critical since a delay of this supplier would not be compensated easily by another supplier within the same business scope. In the figure, three such business scopes can be identified (purple nodes with blue circles), where two of the corresponding suppliers also have a critical score.

\section{Conclusion and Further Research Directions}
\label{sec:conclusion}
Challenges for supply chain management include supply chain intransparency, data disconnectedness and incompleteness, and the scalable identification of criticalities in the supply network. We addressed these challenges by modeling supply chain-related information as a knowledge graph. We used state-of-the-art knowledge graph completion methods to predict missing links and applied graph analysis algorithms to compute importance scores for all suppliers. Based on the importance scores and the graph structure, critical supply chains could be identified, which is an essential step towards more resilient supply networks.

For further research, we propose the following possible directions:
\begin{itemize}
    \item Integration of node and edge properties: In this paper, we only focused on the graph structure for link prediction and graph analysis. For some entity and relation types, however, there exist properties that could be helpful for prediction.
    For example, the relation type \textit{produced\textunderscore in} between a substance and a country has the property Herfindahl-Hirschmann-Index, a measure of market concentration. For the prediction of the relation type \textit{located\textunderscore in}, the company name could be a good indicator. 
    These properties could be integrated when learning embeddings or calculating importance scores. 
    \item Node regression or classification: Besides link prediction, node regression or classification are common tasks on knowledge graphs. Given, e.g., risk scores or categories for a subset of companies, one could learn risk scores or categories for companies that are missing this information in the graph.
    \item Analysis of the complete graph: We conducted the graph analysis based on a subgraph containing the suppliers and the \textit{supplies\textunderscore to} relation type. 
    To calculate the importance scores, more information from the graph could be included. For example, a supplier that is located in a country with a high sustainability risk might also have a higher risk, or a supplier that manufactures many Siemens parts would be more critical for Siemens.  
\end{itemize}


\begin{acknowledgments}
This work has been supported by the German Federal Ministry for Economic Affairs and Climate Action (BMWK) as part of the project CoyPu under grant number 01MK21007K.
\end{acknowledgments}

\bibliography{bibliography}


\end{document}